\documentclass[11pt,a4paper]{article}
\usepackage[hyperref]{emnlp-ijcnlp-2019}
\usepackage{times}
\usepackage{latexsym}
\usepackage{xspace}
\usepackage{url}
\usepackage{amsmath,amsfonts,amssymb}
\usepackage{mathtools}
\usepackage{booktabs}
\usepackage{graphicx}
\usepackage{microtype}
\usepackage{subcaption}

\aclfinalcopy %

\newcommand\papertitle{Unsupervised Discovery of Multimodal Links in Multi-image, Multi-sentence Documents}

\author{Jack Hessel \quad Lillian Lee \quad David Mimno \\
    Cornell University \\
  {\tt \{jhessel, llee\}@cs.cornell.edu \quad mimno@cornell.edu}
}

\title{\papertitle}

\date{}

\newcommand{\suppmat}{\href{http://www.cs.cornell.edu/~jhessel/multiretrieval/multiretrieval.html}{supplementary material}\xspace}

\newcommand{\mparagraph}[1]{\noindent\textbf{{#1}}.}
\newcommand{\mparagraphnp}[1]{\mparagraph{#1}}

\newcommand{\citeposs}[1]{\citeauthor{#1}'s \citeyearpar{#1}}

\newcommand{\aucmath}{{\scriptstyle \sf{AUC}}}
\newcommand{\auc}{$\aucmath$\xspace}

\newcommand{\documenti}{d_i}
\newcommand{\numimages}{m}
\newcommand{\numsents}{n}
\newcommand{\simmat}{M}

\newcommand{\precCutoff}{C}
\newcommand{\patk}{p@\precCutoff} %
\newcommand{\patone}{p@1}

\newcommand{\patfive}{p@5}
\newcommand{\dmm}{d_{{\rm multi}}}
\newcommand{\sentset}{S}

\newcommand{\imageset}{V}

\newcommand{\nnegdoc}{b}
\newcommand{\simf}{{\rm sim}}%
\newcommand{\itpair}[2]{\langle #1, #2\rangle}

\newcommand{\simmati}{\simmat_i}
\newcommand{\sentseti}{\sentset_i}
\newcommand{\imageseti}{\imageset_i}
\newcommand{\numsentsi}{\numsents_i}
\newcommand{\numimagesi}{\numimages_i}

\newcommand{\simdoc}{\simf(\sentset, \imageset)}
\newcommand{\loss}{\mathcal{L}}

\newtoggle{hideOurComments}
\toggletrue{hideOurComments}
\newcommand{\pdfcomment}[3]{\textcolor{#1}{#2: #3}} %

\iftoggle{hideOurComments}{
\renewcommand{\pdfcomment}[3]{}
}

\newcommand{\dii}{Story-DII\xspace}
\newcommand{\sis}{Story-SIS\xspace}
\newcommand{\mscoco}{MSCOCO\xspace}
\newcommand{\diihard}{DII-Stress\xspace}
\newcommand{\rqa}{RQA\xspace}
\newcommand{\diy}{DIY\xspace}
\newcommand{\wiki}{WIKI\xspace}

\newcommand{\DataOne}{Crowdlabeled\xspace}
\newcommand{\dataOne}{crowdlabeled\xspace}

\newcommand{\dataTwo}{organically-multimodal\xspace}

\newif\ifnewresults

\newresultsfalse

\begin{document}
\maketitle
\begin{abstract}
  Images and text co-occur constantly on the web, but explicit links between
images and sentences (or other intra-document textual units) are often not
present. We present algorithms that discover image-sentence relationships
\emph{without} relying on explicit multimodal annotation in training. We
experiment on seven datasets of varying difficulty, ranging from  documents
consisting of groups of images captioned {post hoc} by crowdworkers to
naturally-occurring user-generated multimodal documents. We find that a
structured training objective based on identifying whether \emph{collections}
of images and sentences co-occur in documents can suffice to predict links
between specific sentences and specific images within the \emph{same document}
at test time.
\end{abstract}

\section{Introduction}

Images and text act as natural complements on the modern
web. News stories include photographs, product listings show multiple images
providing detail for online shoppers, and Wikipedia pages include
maps, diagrams, and pictures.
But the exact matching between words and images is often left implicit. %
Algorithms
that identify document-internal connections between specific images and
specific passages of text
could have both immediate and long-term promise.
On the user-experience front,
alt-text for vision-impaired
users could be produced automatically \citep{DBLP:conf/cscw/WuWFS17} via
intra-document retrieval, and user interfaces could explicitly link images to descriptive sentences, potentially improving the
reading experience of sighted users.
Also, in terms of improving other applications,
the text in multimodal documents can be viewed as a noisy
form of image annotation: inferred image-sentence associations can
serve as training pairs for vision models, particularly in domains
lacking readily-available labeled data.
\begin{figure}[t]
  \center
  \includegraphics[width=.85\linewidth]{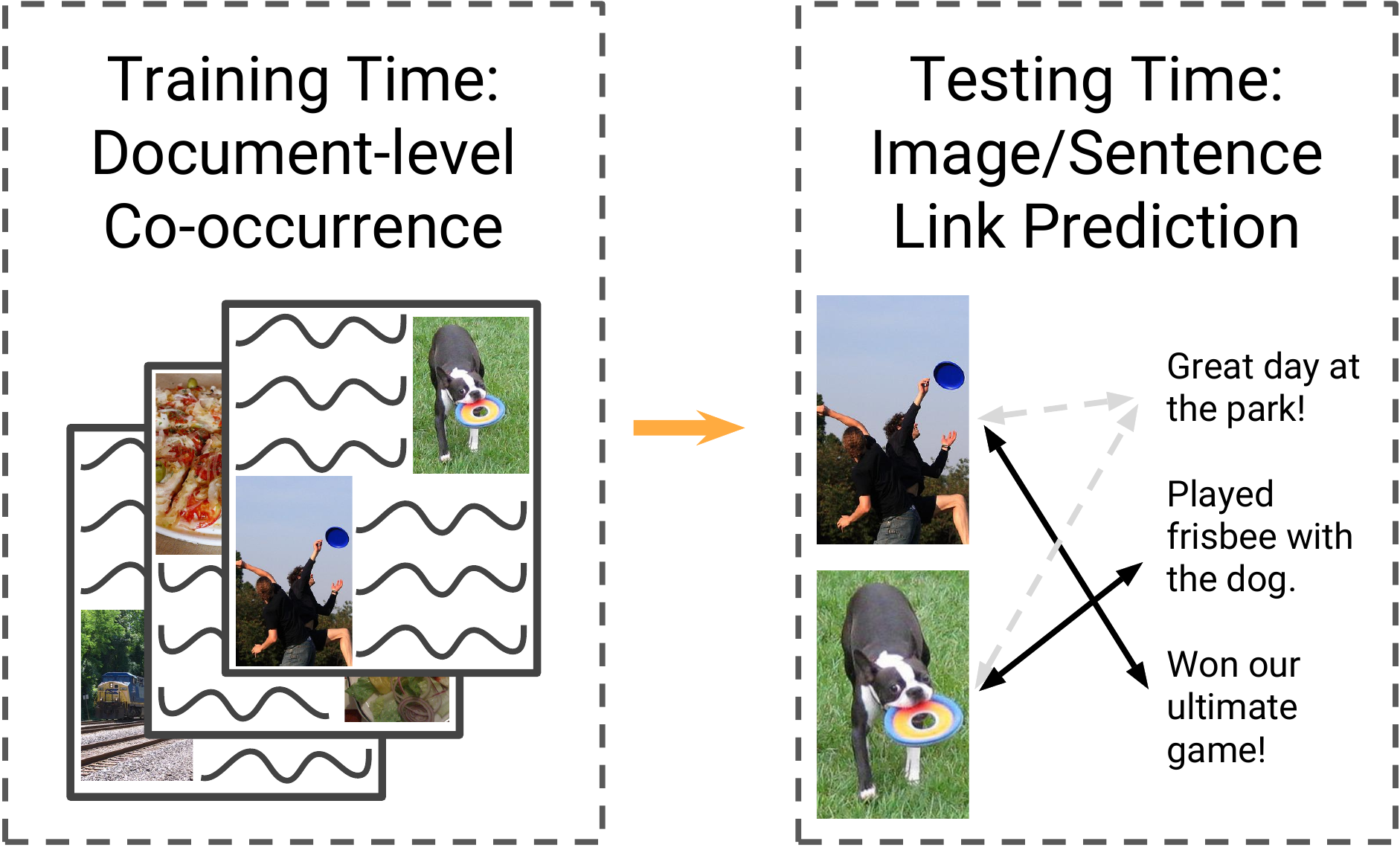}
  \caption{At training time, we assume we are given a set of
    multi-image/multi-sentence documents. At test-time, we predict
    links between individual images and individual sentences within
    single documents. Because no explicit multimodal annotation is
    available at training time, we refer to this task as
    unsupervised.}
  \label{fig:task}
\end{figure}

In this work, we develop \emph{unsupervised} models that learn to identify
multimodal within-document links \emph{despite not
having access to supervision at the individual image/sentence
level during training.}
Rather, the training documents contain multiple images and multiple
sentences\footnote{Any discrete
textual unit
could be used,
such as
n-grams
or paragraphs. We focus on sentences because there
exist
public
sentence-level datasets that we can use for evaluation.} that are not aligned,
as illustrated in Figure~\ref{fig:task}.

Our intra-document setting poses challenges beyond those encountered in the
usual cross-modal retrieval framework, wherein ``documents'' generally consist
of a single image associated with a single
piece
of text, e.g., an image caption.
For the longer documents we consider, a sentence may have many corresponding images or no corresponding images,
and vice versa.  Furthermore, we expect that
images \emph{within} documents will be, on average, more similar than images
\emph{across} documents, thus making disambiguation more difficult than
in the usual one-image/one-sentence case.

Our approach for this difficult setting is ranking-based: we train
algorithms to score image collections and sentence collections that truly co-occur
more highly than image collections and sentence collections that do not
co-occur. The matching functions we consider predict a latent
similarity-weighted bipartite graph over a document's images and
sentences;
at test time,
we evaluate this internal bipartite graph representation learned by our
models for the task of intra-document link prediction.

We work with a variety of datasets (one of which we introduce),
ranging from concatenations of individually-captioned images to
\dataTwo documents scraped from noisy, user-generated web
content.\footnote{Data and
code: \url{www.cs.cornell.edu/\~jhessel/multiretrieval/multiretrieval.html}}
Despite having no supervision at the individual image-sentence level,
our algorithms perform well on the same-document link prediction task.
For example, on a
visual storytelling dataset, we achieve 90+ \auc,
even in the presence of a large number of sentences that do not
correspond to any images in the document. Similarly, for \dataTwo
web data, we are able to surpass object-detection baselines by a
wide margin, e.g., for a step-by-step recipe dataset, we improve
precision by 20 points on link prediction within documents by
leveraging document-level co-occurrence during
training.

We conclude by using our algorithm to
\emph{discover} links within a Wikipedia image/text dataset that lacks
ground-truth image-sentence links. While the predictions are
imperfect, the algorithm qualitatively identifies meaningful patterns,
such as matching an image of a dodo bird to one of two sentences (out of 100)
 in the corresponding article that mention ``dodo''.

\section{Task Formulation}
We assume as given a
set of documents
where each document $\documenti = \langle \sentseti, \imageseti
\rangle$ consists of a set $\sentseti$ of $\numsentsi = |\sentseti|$
sentences and a set $\imageseti$ of $\numimagesi = |\imageseti| $
images.\footnote{Sentences and images can be considered as sequences
  rather than sets in our framework, but unordered sets are more
  appropriate for modeling some of the crowd-sourced corpora we used
  in our experiments.}  For example,
$\documenti$ could
be an article about Paris with $\numsentsi=100$ sentences and
$\numimagesi=3$ images of, respectively,  the Eiffel Tower, the Arc
de Triomphe, and a map of Paris.
For each $\documenti$,
we are to predict
an alignment --- where some sentences or images may not be aligned to anything ---
represented by a (potentially sparse)
bipartite graph on
$\numsentsi$ sentence nodes and $\numimagesi$
image nodes.
During training, we are
\emph{given no access} to ground-truth image-sentence association
graphs, i.e., we do not know \emph{a priori} which images correspond to
which sentences, only that all images/sentences in a document co-occur
together; this is why we refer to our task as \emph{unsupervised}.

We produce a dense sentence-to-image association matrix
$\widehat\simmati \in \mathbb{R}^{\numsentsi \times \numimagesi}$, in
which each entry is the confidence that there is an (undirected) edge
between the corresponding nodes.  Applying different thresholding
strategies to $\widehat\simmati$'s
values yields different
alignment graphs.

\mparagraph{Evaluation} When we have ground-truth alignment graphs for test
documents, we evaluate the correctness of the
association matrix $\widehat\simmati$
predicted by our algorithms according to two metrics:
AUROC (henceforth \auc) and precision-at-$\precCutoff$ ($\patk$). \auc, commonly used in evaluating link
prediction (see \citet{menon2011link})
is the area under the curve of the true-positive/false-positive rate
produced by sweeping over possible confidence thresholds; random is  50, perfect is 100.  $\patk$ measures the accuracy of
the algorithm's most confident $\precCutoff$ predicted edges (in our
case, the most confident edges correspond to the largest entries in
$\widehat\simmati$).
This metric models cases
where only
a small number of high-confidence predictions need be made per document. We
evaluate using $\precCutoff\in \{1,5\}$.

\section{Models}

Our algorithm is inspired by work in cross-modal retrieval
\cite{rasiwasia2010new,hodosh2013framing,pereira2014role,kiros2014unifying}.
Instead of operating at the level of individual images/sentences,
however, our training objective encourages image \emph{sets} and
sentence \emph{sets} appearing in the same document to be more similar
than non-co-occurring sets.

\subsection{Alignment Model and Loss Function}

We assume that the dimensionality $\dmm$ of the multimodal text-image
space is predetermined.

\mparagraphnp{Extracting sentence representations} We pass the words in each sentence
through a 300D word-embedding
layer initialized with GoogleNews-pretrained word2vec embeddings
\cite{mikolov2013distributed}. We then pass the sequence of word vectors
to a GRU \cite{cho2014learning} and extract and L2-normalize a
$\dmm$-dimensional sentence representation
from the final hidden state.

\mparagraphnp{Extracting image representations}
We first compute a
representation for each image using a convolutional neural
network (CNN).\footnote{In some experiments, we use pre-computed image
  features from a pre-trained CNN \citep{sharif2014cnn}. In
  other cases, we fine-tune the full image
  network. We specify which representation we choose in a later
  section.}
The network's output is then mapped via affine projection to $\mathbb{R}^{\dmm}$
and  L2-normalized.

\mparagraphnp{Correspondence prediction}
The result of running the two steps above on
an image-set/text-set pair $\itpair{\sentset}{\imageset}$ is
$|\sentset| +|\imageset|$ vectors, all in $\mathbb{R}^{\dmm}$.
From these, we compute the similarity matrix $\widehat \simmat \in
\mathbb{R}^{|\sentset|\times |\imageset|}$, where the $(j,k)^{th}$
entry is the cosine similarity between the $j^{th}$ sentence vector
and the $k^{th}$ image vector.

\mparagraphnp{Training Objective}
We train
under the assumption that co-occurring image-set/sentence-set pairs should
be more similar than non-co-occurring image-set/sentence-set pairs. We hope
that use of this {\em document-level} objective will produce an
$\widehat \simmati$
offering reasonable {\em intra-document} information at test time, even though such
information is not available at training time.

The training process is modulated by a similarity function $\simdoc$
that measures the similarity between a set of sentences and a set of
images by examining the entries of the individual image/sentence
similarity matrix $\widehat \simmati$ (specific definitions of
$\simdoc$ are proposed in \S \ref{sec:similarity_functions}).
 We use a max-margin loss with negative sampling:
we iterate through true documents $\documenti = \langle
\sentseti, \imageseti \rangle$, and negatively sample at the document
level a set of $\nnegdoc$ sets of images that did not co-occur with
$\sentseti$, $\mathbb{\imageset}'=\{\imageset'_1, ...,
\imageset'_\nnegdoc\}$, and a set of $\nnegdoc$ sets of sentences that
did not co-occur with $\imageseti$, $\mathbb{\sentset}'=\{\sentset'_1,
..., \sentset'_\nnegdoc\}$.

We then compute a loss for $\langle \sentseti, \imageseti \rangle$ by
comparing the true similarities to the negative-sample similarities.
We find that hard-negative mining
\cite{dalal2005histograms,schroff2015facenet,faghri2018vse++}, the
technique of selecting the negative cases that maximally violate the
margin within the minibatch, performs better than simple averaging. The
loss for a single positive example is:
\begin{equation}
  \begin{split}
    \loss\left(\sentseti, \imageseti \right) = \max_{\imageset' \in \mathbb{\imageset}'}
    h\left(\simf(\sentseti, \imageseti), \simf(\sentseti, \imageset')\right)\\
    + \max_{\sentset' \in \mathbb{\sentset}'}
    h\left(\simf(\sentseti, \imageseti), \simf(\sentset', \imageseti)\right)
  \end{split}
  \label{eq:loss_function}
\end{equation}
for hinge loss $h_\alpha(p,n)={\max(0, \alpha-p+n)}$, where we set
margin $\alpha=0.2$ \citep{kiros2014multimodal,faghri2018vse++}.

\subsection{Similarity Functions}
\label{sec:similarity_functions}

We explore several functions for measuring how similar a set of
$\numsents$ sentences $\sentset$ is to a set of $\numimages$ images
$\imageset$. All similarity functions convert the matrix
$\widehat{\simmat} \in \mathbb{R}^{\numsents\times \numimages}$
corresponding to $\itpair{\sentset}{\imageset}$ into a
bipartite graph based on the magnitude of the entries.
The functions differ in how they determine which entries
$\widehat{\simmat_{ij}}$ correspond to edges and edge weights.

\mparagraph{Dense Correspondence (DC)} %
The DC function assumes a dense correspondence between images and sentences;
each sentence must be aligned to its
most similar image, and vice versa, regardless of how small the similarity
might be:
\begin{equation*}
  \begin{split}
  \simdoc =  \frac{1}{\numsents} \sum_{i=0}^\numsents \max_{j} \widehat M_{i,j}
  + \frac{1}{\numimages} \sum_{j=0}^\numimages \max_{i} \widehat M_{i,j}.
  \end{split}
  \label{eq:dc_sim}
\end{equation*}
The underlying assumption of this function can clearly be violated
in practice:\footnote{\citet[\S3.3.1]{karpathy2014deep} discuss violations in the image fragment/single-word case.} sentences can have no image, and images no sentence.

\mparagraph{Top-K (TK)} Instead of assuming that every sentence has a
corresponding image and vice versa, in this function only the
top $k$ most likely sentence $\Rightarrow$ image (and image
$\Rightarrow$ sentence) edges are aligned. This process mitigates the effect of non-visual sentences by allowing
algorithms to align them to no image.
We discuss choices of $k$ for particular experimental settings in
\S\ref{sec:crowd-results}.

\begin{figure*}[ht]
  \centering
  \includegraphics[width=.99\linewidth]{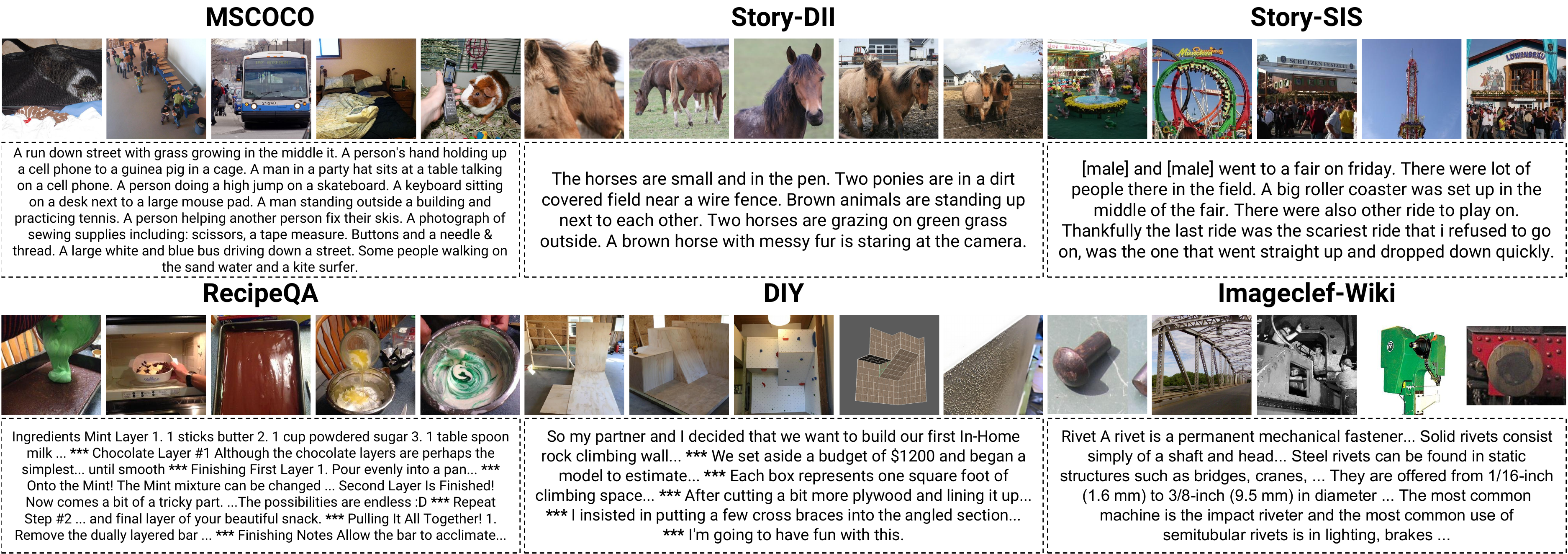}
  \caption{Sample documents from six of our datasets. Image sets and
    sentence sets may be truncated due to space constraints.
    The example from \dii is harder than is typical, but we include it to illustrate
    a point regarding image spread made in \S \ref{sec:crowd-results}.
    *** denotes text-chunk
    delimiters present in the original data.
    }
  \label{fig:datasets}
\end{figure*}

\label{sec:section_with_assignment_problem}
\mparagraph{Assignment Problem (AP)}
We may wish to consider the image-sentence alignment task as a \emph{bipartite linear
assignment problem} \citep{kuhn1955hungarian}, such that each
image/sentence in a document has at most one association. Each time we
compute $\simdoc$ in the forward pass of our models, we solve the
integer programming problem of maximizing $\sum_{i,j} \widehat \simmat_{ij} x_{ij}$
subject to the constraints:
\begin{equation*}
\textcolor{red}{\forall i, \sum_j x_{ij} \leq  1;}\,  \textcolor{blue}{\forall j, \sum_i x_{ij} \leq 1;}\,  \textcolor{magenta}{\forall i, j, x_{ij} \in \{0, 1\}.}
\end{equation*}
\label{eq:assignment}
Despite involving a discrete
optimization step, the model remains fully differentiable. Our forward
pass uses tensorflow's python interface,
\texttt{tf.py\_func}, and the \texttt{lapjv} implementation of
the JV algorithm \cite{jonker1987shortest} to solve the integer
program itself. Given the solution $x^*_{ij}$,
we compute (and backpropagate gradients through) the similarity function
$\simdoc = \left( \sum_{i,j} \simmat_{ij}x^*_{ij} \right) / r$ where
$r$ is the number of non-zero $x^*_{ij}$. 
Should we want to impose an upper bound $k$ on the
number of links, we can add the following additional
constraint:\footnote{Applying \citeposs{volgenant2004solving}
  polynomial-time algorithm.}
$\sum_{i,j} x_{ij} \leq k(\sentset, \imageset)$.  For example, one could
set $k(\sentset, \imageset) = \frac{1}{2} \min(|\sentset|,
|\imageset|)$.

The JV algorithm's runtime is $O(\max(n, m)^3)$, and each positive example
requires computing similarities for
the positive case and the $2\nnegdoc$ negative samples from
Eq.~\ref{eq:loss_function}, for a per-example runtime of $O(b
\cdot \max(n,m)^3)$. Fortunately, \texttt{lapjv} is highly optimized,
so despite solving many integer programs,
AP often runs \emph{faster}
than DC.

\subsection{Baselines}

We construct two baseline similarity functions, as we are not aware of existing models
that directly address our task in an unsupervised fashion.

\mparagraph{Object Detection} For each image in the document, we use
DenseNet169 \citep{huang2017densely} to find its $K$ most probable ImageNet classes (e.g., ``stingray''),
 and represent the image as the average of the word2vec embeddings
of those $K$  labels.
We represent each
sentence in a document as the mean word2vec embedding of its words. To
form the strongest possible baseline, we compute the cosine similarity
between all sentence-image pairs to form $\widehat \simmat$ for $K\in
\{1 ... 20\}$ and report the variant with the {\em best post-hoc performance on
the test set}.

\mparagraph{NoStruct} The similarity functions described in
\S\ref{sec:similarity_functions} rely on document-level, structural
information, i.e., for a single image in a document, the \emph{other
  images} in a document affect the overall similarity (and vice versa
for sentences). However, this structural information may not be worth
incorporating. Thus, we train a baseline that solely relies on single
image/single sentence co-occurrence statistics. At training time, we
randomly sample a single image and a single sentence from a document,
compute the cosine similarity of their vector representations, and
treat that value as the document similarity. While the randomly
sampled image/sentence will not truly correspond for every sample, we
still expect this baseline to produce above-random results when
averaged over many iterations, as true correspondences have some (low)
probability of being sampled.\footnote{This probability is equal to
  the density of the ground-truth, underlying image-sentence
  association graph.}

\section{
  Experiments on
 \DataOne Data
}
\label{sec:dataone}

Our first set of experiments uses four pre-existing datasets
created by asking crowdworkers to add sentence-long textual descriptions to
images in a collection.  Image-sentence
alignments are therefore known by construction. We do not use these labels
at training time: gold-standard alignments are only  used at evaluation
time to compare performance between
algorithms.\footnote{The \suppmat gives more details.} Statistics of these datasets are given in
the top half of Table~\ref{tab:data_stats}, and example documents are
given in Figure~\ref{fig:datasets}. Each \dataOne dataset is
constructed to address a different question about our learning
setting.

\mparagraph{Q: Is this task even possible? Test: \mscoco} MSCOCO
\citep{lin2014microsoft} was created by crowdsourced manual captioning
of single images.  We construct ``documents'' from this data by first
randomly aggregating five image-caption pairs. We then add five
``distractor'' images with no captions and five ``distractor''
captions with no images. Thus, a non-distractor image truly corresponds to the
single caption that was written about it, and not to the other 9
captions in the document.  There are a total of 10 images/sentences
per document, and 5 ground-truth image-sentence links. \emph{A
  priori,} we expect this to be the easiest setting for
within-document disambiguation because mismatched images and sentences
are completely independent.

\mparagraph{Q: What if the images/sentences within a document are similar?
  Test: \dii} \citet{huang2016visual} asked crowdworkers to collect
subsets of images contained in the same Flickr album
\citep{DBLP:journals/cacm/ThomeeSFENPBL16} that could be arranged into
a visual story.  In the \dii (= ``\underline{d}escriptions
\underline{i}n \underline{i}solation'') case, (possibly different)
crowdworkers subsequently captioned the images, but only saw each
image in isolation.
We construct a set of documents from \dii so that
each contains five images and five
sentences.
Because images come from the same album, images and captions in
our \dii ``documents'' are %
more similar to each other than those
in our MSCOCO ``documents.''

\mparagraph{Q: What if the sentences are cohesive and refer to
  each other? Test: \sis} \newcite{huang2016visual} also presented all
the images in a subset from the same Flickr album to crowdworkers
simultaneously and asked them to caption the image subsets
collectively to form a story (SIS = ``\underline{s}tory \underline{i}n
\underline{s}equence'').  In contrast to \dii, the generated sentences
are
generally not
stand-alone descriptions of the corresponding image's contents, and may,
for example, use pronouns to refer to elements from neighboring sentences and images.

\mparagraph{Q: What if there are many
sentences with no
  corresponding images? Test: \diihard}
Because documents often have many sentences that do not directly refer to visual content,
we constructed a setting with many more
sentences than images.
We augment documents from \dii with 45 randomly negatively sampled
distractor captions. The resulting documents have five images and
fifty sentences, where only five sentences truly describe images in
the document.

\begin{table}
  \small
  \centering
  
\begin{tabular}{l@{\hspace{.2cm}}c@{\hspace{.2cm}}c@{\hspace{.2cm}}r@{\hspace{.2cm}}r}
\toprule
& train/val/test & $\numsentsi/\numimagesi$ & \# imgs & density \\
&                &  (median)                       & (unique) & \\
\midrule
\mscoco & 25K/2K/2K & 10/10 & 83K & 5\% \\
\dii & 22K/3K/3K & 5/5 & 47K & 20\% \\
\sis & 37K/5K/5K & 5/5 & 76K & 20\% \\
\diihard & 22K/3K/3K & 50/5 & 47K & 2\% \\
\midrule
\diy & 7K/1K/1K & 15/16 & 154K & 8\% \\
\rqa & 7K/1K/1K & 6/8 & 88K & 17\% \\
\wiki & 14K/1K/1K & 86/5 & 92K & N/A \\
\bottomrule
\end{tabular}

  \caption{Dataset statistics: top half = \dataOne datasets;
    bottom half = \dataTwo datasets. {\em Density} measures the
     sparsity of the ground truth graph as the number of ground-truth
    edges divided by the number of possible edges.}
  \label{tab:data_stats}
\end{table}

\begin{figure*}
  \centering
  \begin{minipage}{.68\textwidth}
    \small \centering 
\begin{tabular}{lc@{\hspace{.1cm}}c@{\hspace{.25cm}}c@{\hspace{.1cm}}c@{\hspace{.25cm}}c@{\hspace{.1cm}}c@{\hspace{.25cm}}c@{\hspace{.1cm}}c@{\hspace{0cm}}}
\toprule
&\multicolumn{2}{c}{\mscoco}&\multicolumn{2}{c}{\dii}&\multicolumn{2}{c}{\sis}&\multicolumn{2}{c}{\diihard}\\
& \auc & \patone/\patfive & \auc & \patone/\patfive & \auc & \patone/\patfive & \auc & \patone/\patfive \\
 \midrule
Random& 49.7 & 5.0/4.6 & 49.4 & 19.5/19.2 & 50.0 & 19.4/19.7 & 50.0 & 2.0/2.0  \\
Obj Detect& 89.5 & 67.7/45.9 & 65.3 & 50.2/35.2 & 58.4 & 40.8/28.6 & 76.9 & 25.7/17.5  \\
NoStruct& 87.5 & 50.6/34.6 & 76.6 & 60.1/46.2 & 64.9 & 43.2/33.7 & 84.2 & 21.4/15.6  \\
\midrule
DC & \textbf{98.9} & 93.6/\textbf{80.1} & \textbf{82.8} & 71.5/\textbf{55.5} & \textbf{68.8} & \textbf{51.8}/\textbf{38.6} & \textbf{94.9} & \textbf{64.6}/\textbf{44.8}  \\
\midrule
TK & \textbf{98.9} & 93.9/\textbf{80.1} & \textbf{82.9} & 71.4/\textbf{55.5} & \textbf{68.8} & 50.9/\textbf{38.7} & \textbf{95.2} & \textbf{65.6}/\textbf{45.3}  \\
\, \rotatebox[origin=c]{180}{$\Lsh$} + $\frac{1}{2}k$& \textbf{99.0} & \textbf{95.0}/\textbf{81.1} & \textbf{82.0} & \textbf{72.6}/\textbf{54.9} & 67.6 & \textbf{51.9}/\textbf{38.0} & \textbf{94.7} & 64.0/43.7  \\
\midrule
AP & \textbf{98.7} & 91.0/78.0 & \textbf{82.6} & 70.5/\textbf{55.0} & \textbf{68.5} & 50.5/\textbf{38.3} & \textbf{95.3} & \textbf{65.5}/\textbf{45.7}  \\
\, \rotatebox[origin=c]{180}{$\Lsh$} + $\frac{1}{2}k$& \textbf{98.9} & 93.9/\textbf{80.4} & 81.6 & \textbf{72.4}/54.4 & 67.4 & \textbf{52.1}/37.7 & \textbf{94.5} & \textbf{65.0}/43.4  \\
\bottomrule
\end{tabular}

    \captionof{table}{
      Results for \dataOne datasets
      (similar results for other settings are
      included in the \suppmat). Values are bolded if
      they are within 1\% of the best-in-column performance.}
    \label{tab:ground_truth_res}
  \end{minipage}
  \hfill
  \begin{minipage}{.3\textwidth}
    \centering
    \quad Number of Epochs
      \begin{subfigure}{\textwidth}
        \centering
        \includegraphics[width=.8\linewidth]{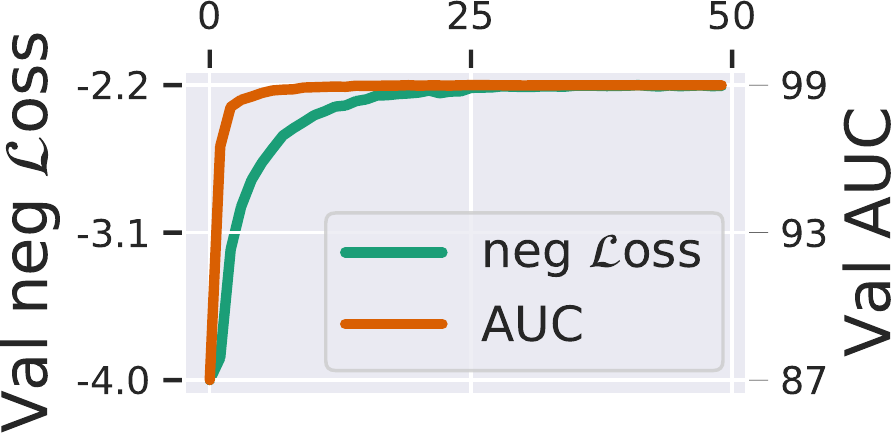}
        \caption{\mscoco}
        \label{fig:coco_dynamics}
      \end{subfigure}
      \begin{subfigure}{\textwidth}
        \centering
        \includegraphics[width=.8\linewidth]{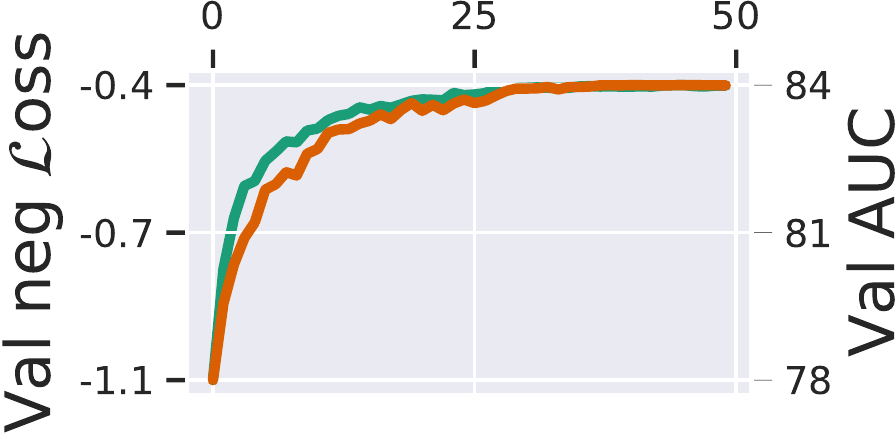}
        \caption{\dii}
        \label{fig:dii_dynamics}
      \end{subfigure}
      \caption{Inter-document objective (AP, $\nnegdoc=10$) and
        intra-document \auc increase together during training.
      }
      \label{fig:outcomes}
  \end{minipage}
\end{figure*}

\mparagraph{Experiment Protocols}
We conduct our
evaluations over a single randomly sampled train/dev/test split.
For image features, we extract the
pre-classification layer of DenseNet169 \cite{huang2017densely}
pretrained on the ImageNet \cite{russakovsky2015imagenet}
classification task, unless otherwise specified. We train with Adam
\cite{kingma2014adam} using a starting learning rate of .0001
for 50
epochs.
We decrease the learning rate by a factor of 5 each time
the loss in Eq.~\ref{eq:loss_function} over the dev set plateaus
for more than 3 epochs.
We set\footnote{
Anecdotally, we found
  that values of 256 and 512 produced similar performance in early testing.
} %
 $\dmm=1024$, and apply dropout with $p=.4$. At test time, we use the
model checkpoint with the lowest dev error.

\subsection{\DataOne-Data Results}
\label{sec:crowd-results}

We tried all combinations of $\nnegdoc \in \{10,20,30\}$, $\simf \in
\{{\rm DC}, {\rm TK}, {\rm AP}\}$. %
For TK and AP we set the maximum
link threshold $k$ to $\min(\sentseti, \imageseti)$ or $\lceil \frac{1}{2}\min(\sentseti, \imageseti)
\rceil$ (denoted $\frac{1}{2}k$ in the results table).\footnote{
  For datasets where $\numimagesi=\numsentsi$ and the first choice
  of definition for $k$ is used,
  $DC$ and $TK$ are the same.
  But running the duplicate
  algorithms anyway provides us with a rough sense of run-to-run
  variability.}

Table~\ref{tab:ground_truth_res} shows test-set prediction results for
$\nnegdoc=10$ (results for $\nnegdoc \in \{20,30\}$ are similar). The
retrieval-style objectives we consider encourage algorithms to learn
useful within-document representations, and incorporating a structured
similarity is beneficial. All our algorithms outperform the strongest
baseline (NoStruct) in all cases, e.g., by
at least
10 absolute percentage
points in \patone\xspace on \dii.

We next show,
as a sanity check,
that our inter-document training objective function (Eq.~\ref{eq:loss_function})
corresponds to intra-document prediction performance (the actual function of interest).
Figure~\ref{fig:outcomes} plots
how both functions vary with number of epochs, for two
different validation datasets.
In general, inter-document performance and intra-document
performance rise together during training;\footnote{See the
  \suppmat for
  plots for all datasets; while the
  general pattern is the same, some of the training curves exhibit
  additional interesting patterns.} for a fixed neural architecture,
models better at optimizing the inter-document loss in
Eq.~\ref{eq:loss_function} also generally produce better
intra-document representations.

In addition, we found that i) DC, despite assuming every sentence
corresponds to an image, achieves high performance on \diihard, even
though 90\% of its sentences do not correspond to an image; ii) Allowing
AP/TK to make fewer connections (i.e., setting $\frac{1}{2}k$) did not
result in significant performance changes, even in the \mscoco case,
where the true number of links (5) was the same as the number of links
accounted for by AP/TK+$\frac{1}{2}k$; and iii) adding topical cohesion
(\mscoco $\rightarrow$ \dii) makes the task more difficult,
as does adding textual cohesion (\dii $\rightarrow$ \sis).

\mparagraph{Models have trouble with the same documents} We calculated
\auc for each test document individually. The Spearman correlation
between these individual-instance \auc values is very high: of all
pairs in DC/TK/AP, over all \dataOne datasets at b=10, DC vs.~AP on
MSCOCO had the lowest correlation with $\rho=.89$.

\label{sec:content_spread}
\mparagraph{Error analysis: content vs. spread} Why are some instances
more difficult to solve for all of our algorithms? We consider two
hypotheses. The ``content'' hypothesis is that some concepts are more
difficult for algorithms to find multimodal relationships between:
``beauty'' may be hard to visualize, whereas ``dog'' is a concrete
concept
\cite{lu2008high,berg2010automatic,parikh2011interactively,hessel2018quantifying,mahajan2018exploring}.
The ``spread'' hypothesis, which we introduce, is that documents with
lower diversity among images/sentences may be harder to disambiguate
at test time. For example, a document in which all images and all
sentences are about horses requires finer-grained distinctions than a
document with a horse, a barn, and a tractor.
The \dii vs. \sis example in Fig.~\ref{fig:datasets} illustrates this
contrast.

To quantify the spread of a document, we first extract vector
representations of each test image/sentence.\footnote{We use
  DenseNet169 features for images and mean word2vec for sentences. We
  don't use internal model representations as we aim to quantify
  aspects of the dataset itself.}  We then L2-normalize the vectors and
compute the mean squared distance to their centroid;
higher ``spread'' values indicate that a document's sentences/images
are more diverse. To quantify the content of a document, for
simplicity, we mean-pool the image/sentence representations
and reduce to 20 dimensions with PCA.

We first compute an OLS regression of image spread + text spread on
test \auc scores for \dii/\sis/\diihard\footnote{\mscoco is omitted
  because the \auc scores are all large.} for AP with $\nnegdoc=10$:
42/23/16\% respectively (F-test $p \ll .01$) of the
variance in \auc can be explained by the spread hypothesis alone. In
general, documents with less diverse content are harder, with image
spread explaining more variance than text spread. When adding in the
image+text content features, the proportion of \auc variance  explained
increases to 52/35/38\%; thus, for these datasets, both the
``content'' and ``spread'' hypotheses independently explain document
difficulty, though the relative importance of each varies
across datasets.

\section{
Experiments on
RQA and DIY
}
\label{sec:web-results}

The previous datasets had captions added by crowdworkers for the
explicit purpose of aiding research on grounding: for MSCOCO, annotators
providing image captions were explicitly instructed to provide literal
descriptions and ``not describe what a person might say''
\cite{chen2015microsoft}.
The manner in which users interact with
multimodal content ``in the wild'' significantly differs from \dataOne
data:
\citeposs{marsh2003taxonomy} 49-element taxonomy of multimodal relationships (e.g., ``decorate'', ``reiterate'',
``humanize'') observed in 45 web documents highlights the diversity of
possible image-text relationships.

We thus consider
two datasets (one of which we release ourselves) of \dataTwo documents scraped from web data, where
the original authors created or selected both images and sentences.
Statistics of these datasets are given in the bottom half
of Table~\ref{tab:data_stats}.

\begin{table}[t]
  \small
  \centering
  
\begin{tabular}{lc@{\hspace{.1cm}}cc@{\hspace{.1cm}}c}
\toprule
&\multicolumn{2}{c}{\rqa}&\multicolumn{2}{c}{\diy}\\
& \auc & \patone/\patfive & \auc & \patone/\patfive \\
 \midrule
Random& 49.4 & 17.8/16.7 & 49.8 & 6.3/6.8  \\
Obj Detect& 58.7 & 25.1/21.5 & 53.4 & 17.9/11.8  \\
NoStruct& 60.5 & 33.8/27.0 & 57.0 & 13.3/11.8  \\
\midrule
DC & 63.5 & 38.3/30.6 & 59.3 & 20.8/16.1  \\
\midrule
TK & 67.9 & 44.0/35.8 & 60.5 & 21.2/16.0  \\
 \,  \rotatebox[origin=c]{180}{$\Lsh$} + $\frac{1}{2}k$& 68.1 & 44.5/35.4 & 56.0 & 14.1/12.5  \\
\midrule
AP & \textbf{69.3} & \textbf{47.3}/\textbf{37.3} & \textbf{61.8} & \textbf{22.5}/\textbf{17.2}  \\
\, \rotatebox[origin=c]{180}{$\Lsh$} + $\frac{1}{2}k$& \textbf{68.7} & \textbf{47.2}/36.2 & 59.4 & \textbf{21.6}/15.3  \\
\bottomrule
\end{tabular}

  \caption{Performance on the \dataTwo data; values
    within 1\% of best-in-column are bolded.}
  \label{tab:webdata_results}
\end{table}

\begin{figure*}[t!]
  \centering
  \begin{subfigure}[t]{0.5\linewidth}
    \centering
    \includegraphics[width=.98\linewidth]{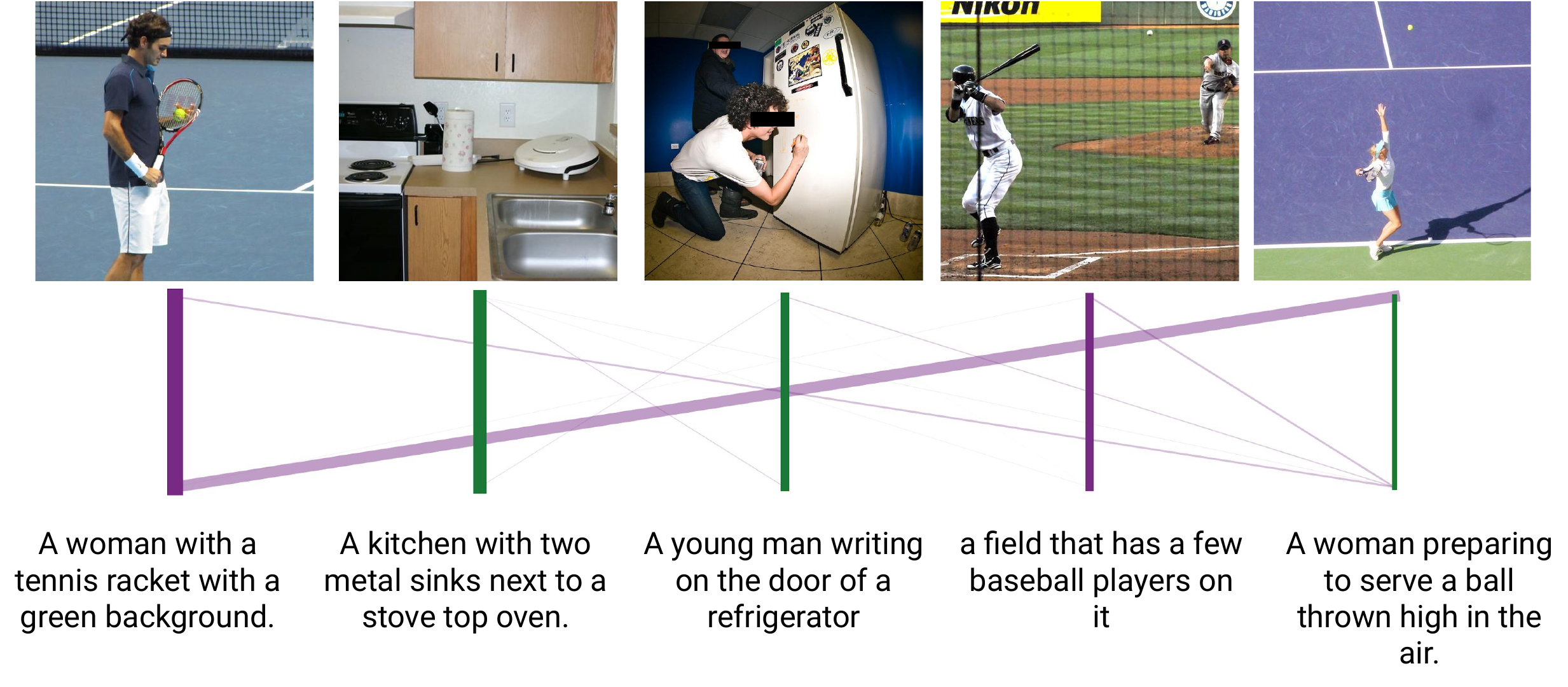}
    \caption{\mscoco; 97 \auc, 10 sentences/10 images.}
    \label{fig:example_prediction_mscoco}
  \end{subfigure}~
  \begin{subfigure}[t]{0.5\linewidth}
    \centering
    \includegraphics[width=.98\linewidth]{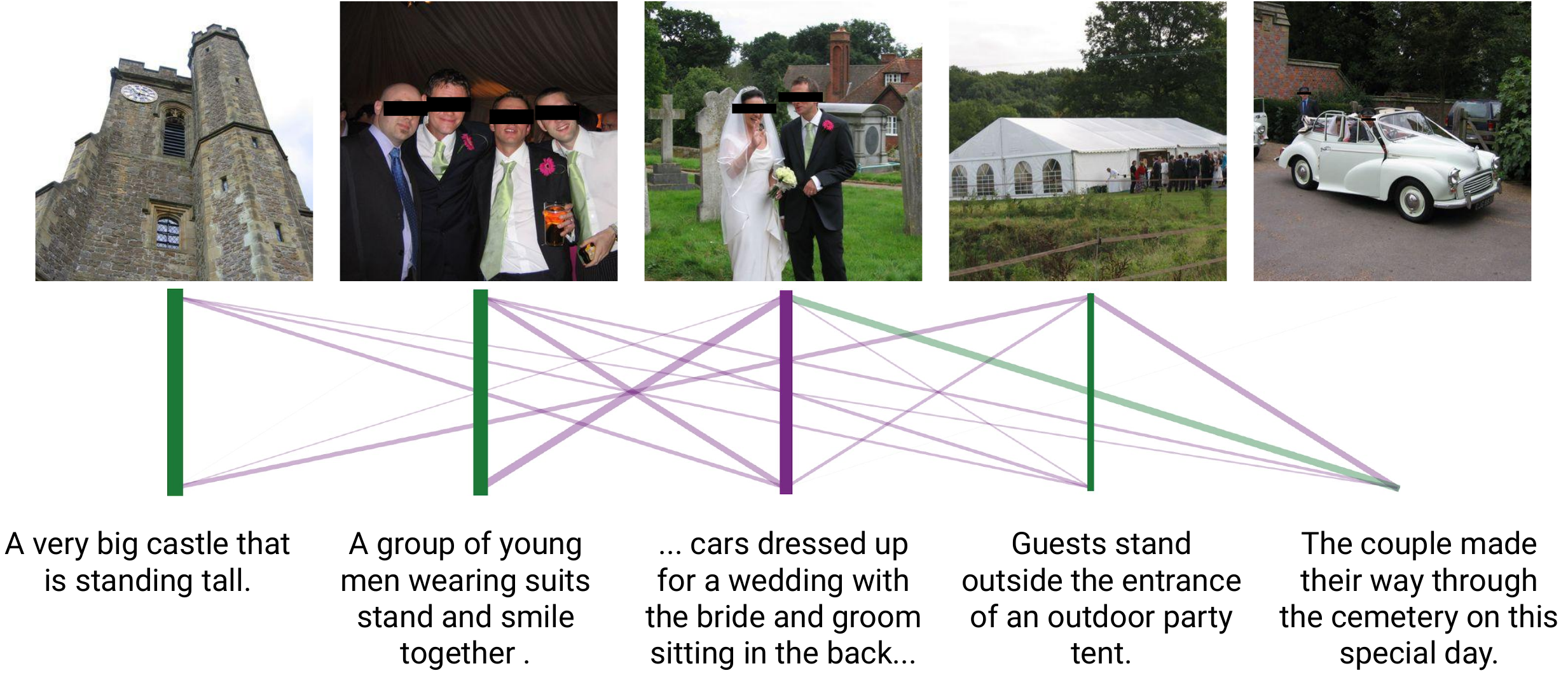}
    \caption{\dii; 83 \auc, 5 sentences/5 images.}
    \label{fig:example_prediction_dii}
  \end{subfigure}
  \begin{subfigure}[t]{0.5\linewidth}
    \centering
    \includegraphics[width=.98\linewidth]{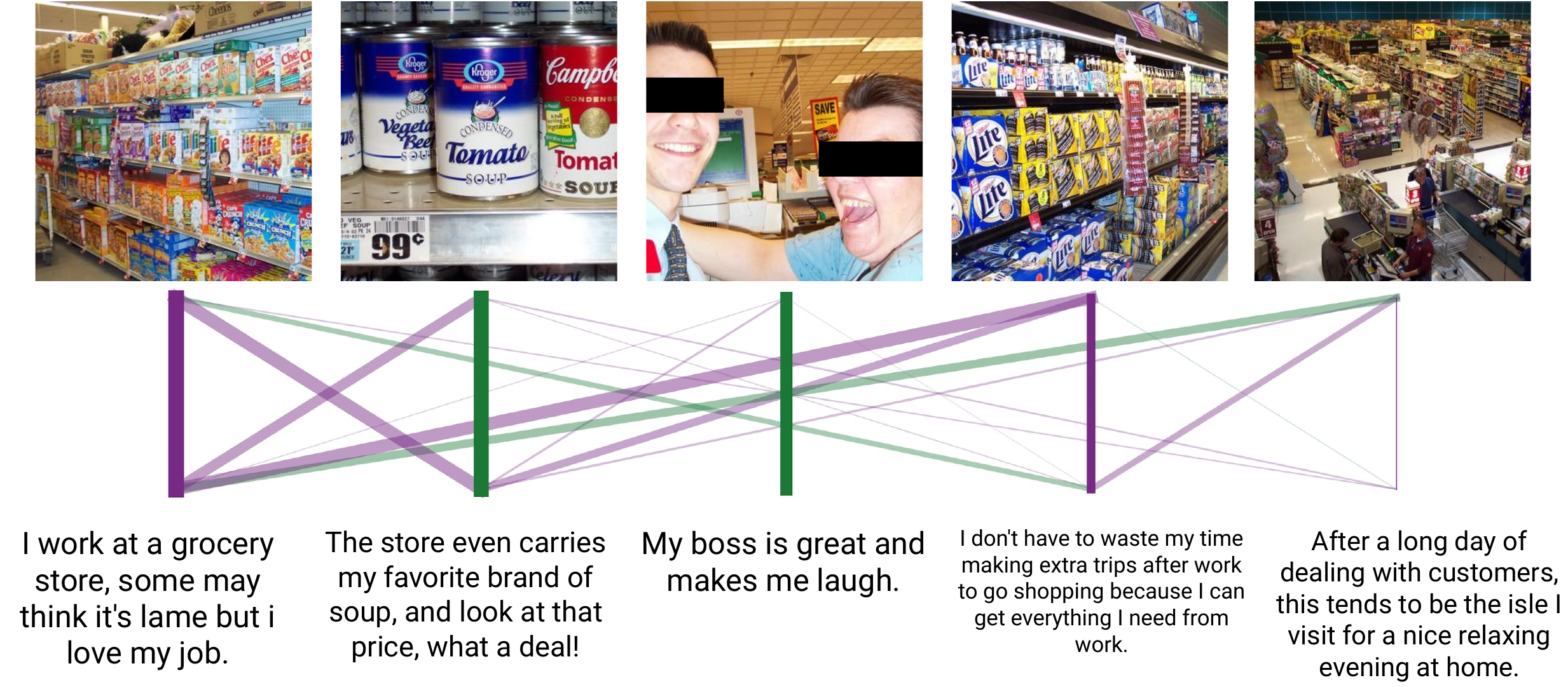}
    \caption{\sis; 70 \auc, 5 sentences/5 images.}
    \label{fig:example_prediction_sis}
  \end{subfigure}~
  \begin{subfigure}[t]{0.5\linewidth}
    \centering
    \includegraphics[width=.98\linewidth]{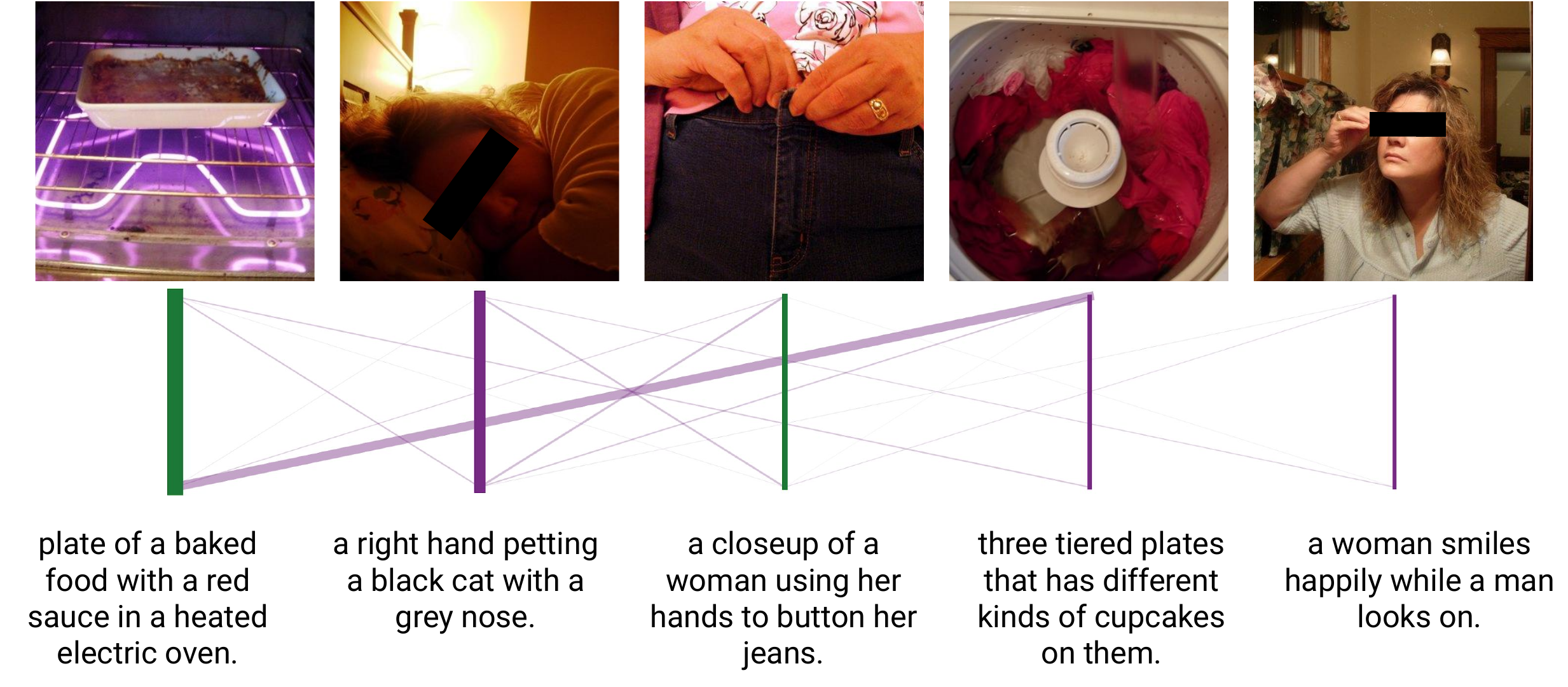}
    \caption{\diihard; 94 \auc, 50 sentences/5 images.}
    \label{fig:example_prediction_diir}
  \end{subfigure}
  \begin{subfigure}[t]{0.5\linewidth}
    \centering
    \includegraphics[width=.98\linewidth]{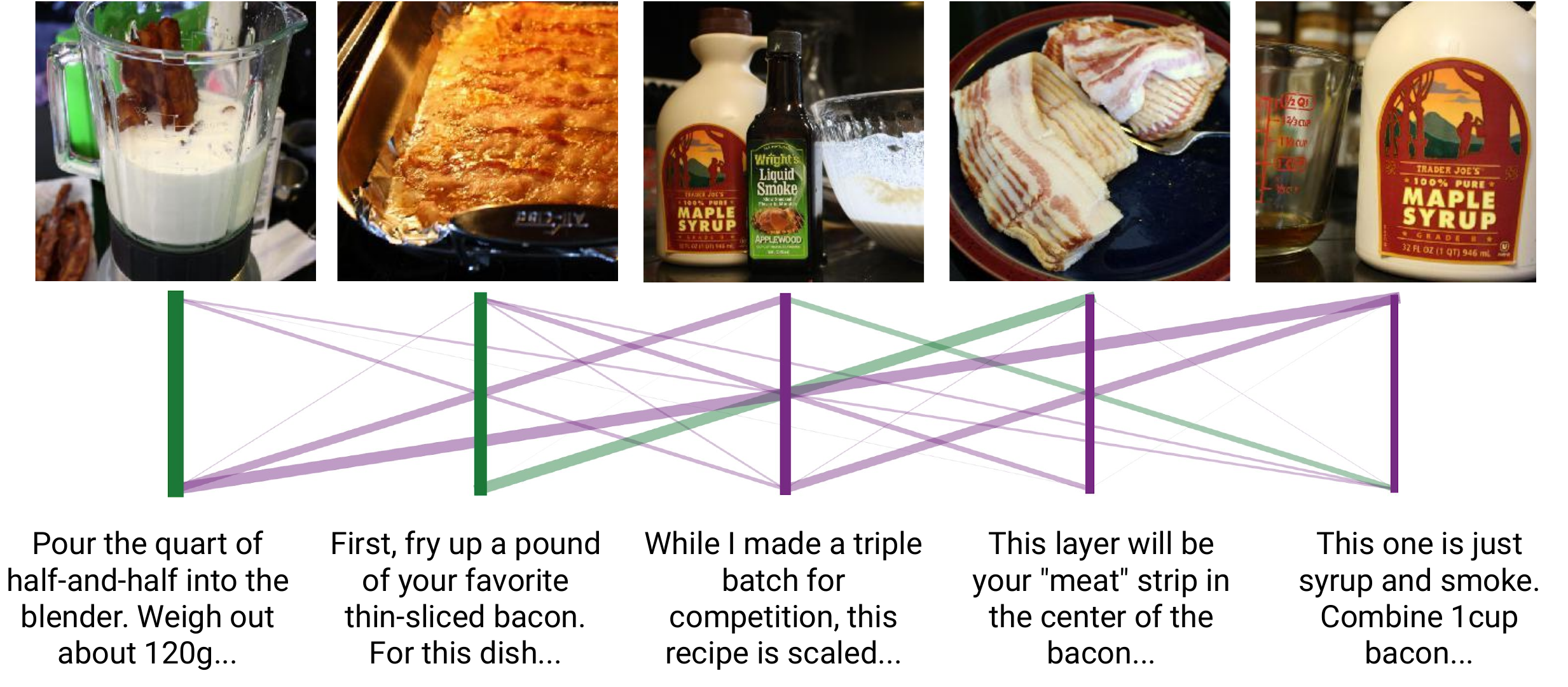}
    \caption{\rqa; 70 \auc, 9 sentences/18 images.}
    \label{fig:example_prediction_rqa}
  \end{subfigure}~
  \begin{subfigure}[t]{0.5\linewidth}
    \centering
    \includegraphics[width=.98\linewidth]{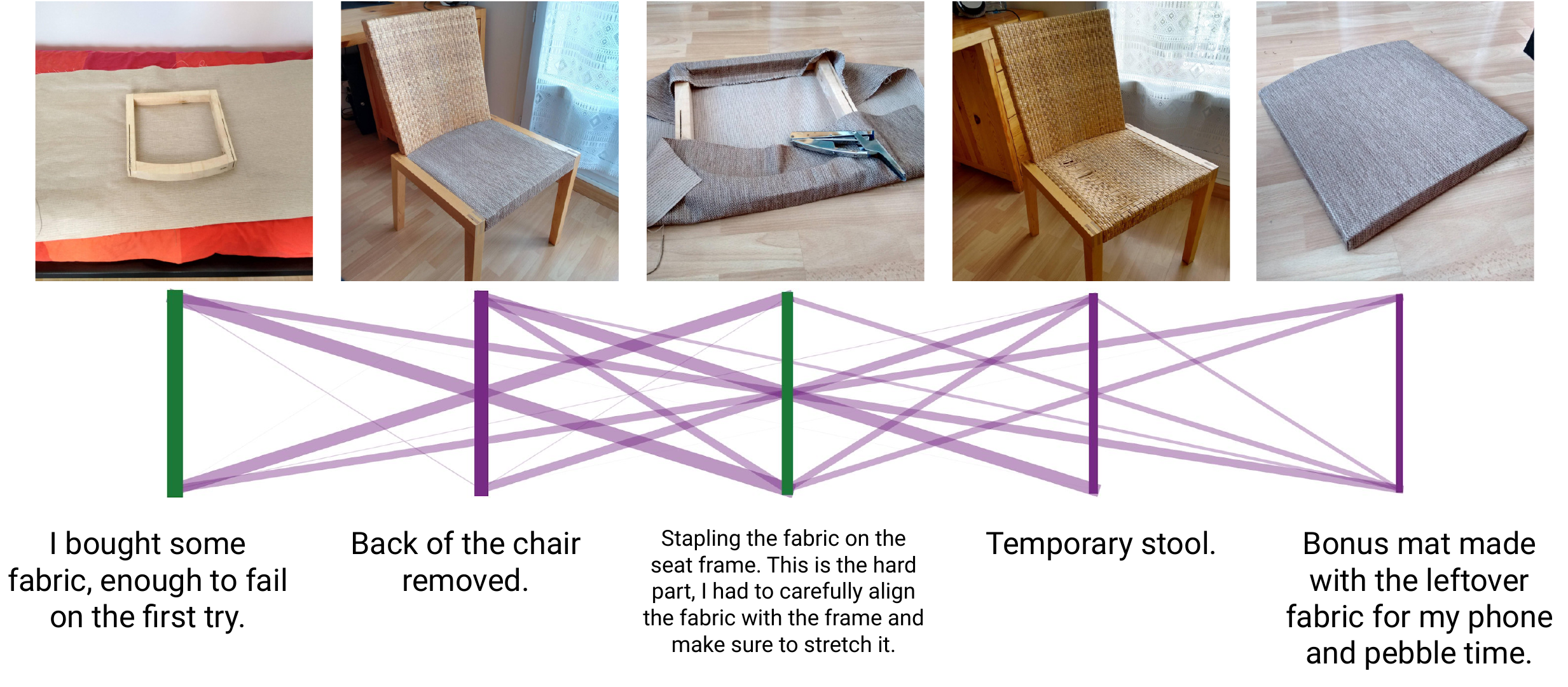}
    \caption{\diy; 62 \auc, 17 sentences/17 images.}
    \label{fig:example_prediction_diy}
  \end{subfigure}
  \caption{Example test-time graph predictions from AP with $b=10$. Each subfigure
  gives the top 5 image/sentence predictions per document, in decreasing order of confidence from left to right.
 Green edges indicate ground-truth pairs;
edge widths show the magnitude of edges in $\widehat\simmati$ (only positive weights are shown).
 Examples are selected to be representative: per-document \auc (roughly) matches the average \auc achieved on the corresponding dataset.}
  \label{fig:example_predictions_all}
\end{figure*}

\mparagraph{\rqa} RecipeQA \cite{yagcioglu2018recipeqa} is a
question-answering dataset scraped from \texttt{instructibles.com}
consisting of images/descriptions of food preparation steps;
we construct documents by treating each recipe step as
a sentence.\footnote{Recipe steps have variable length, are often not
  strictly grammatical sentences, and can contain lists, linebreaks,
  etc.} Users of the Instructibles web interface put images and recipe steps
in direct correspondence, which gives us a graph for test time evaluation.

\begin{figure*}
  \centering
  \includegraphics[width=.95\linewidth]{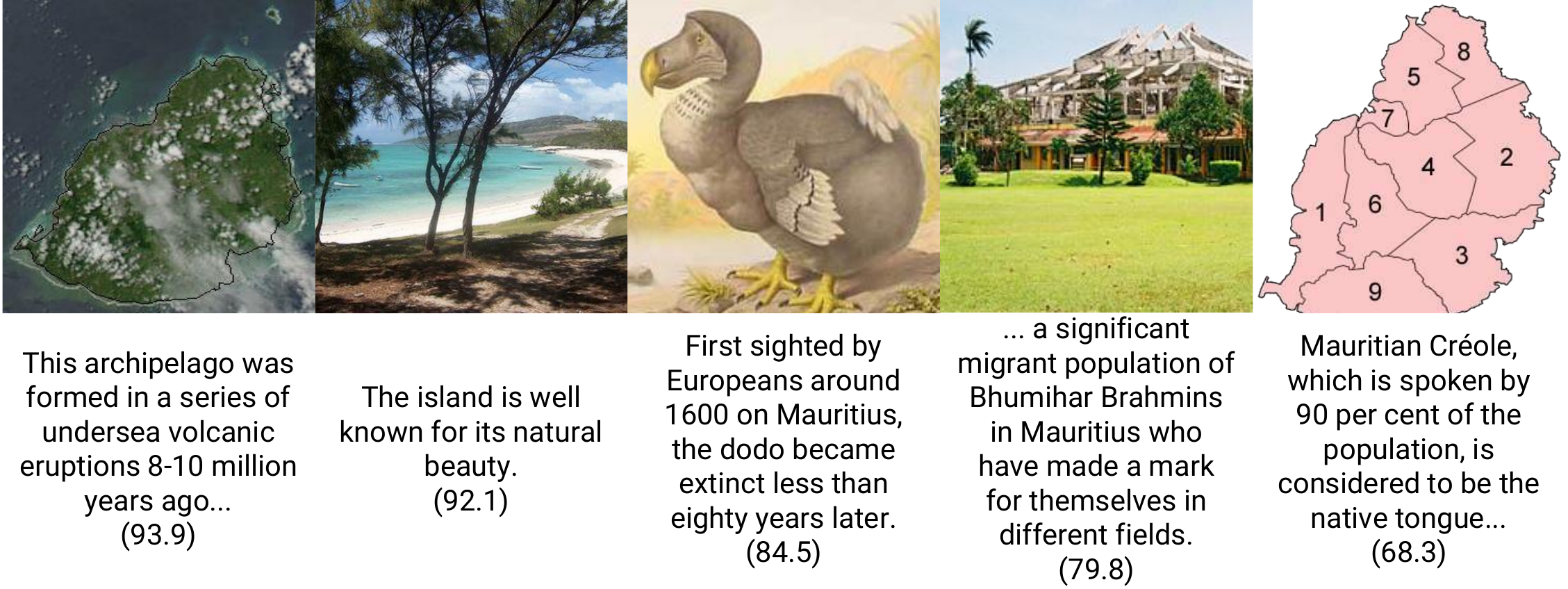}
  \caption{Predicted sentences, with cosine similarities, for
    images in a 100-sentence
    ImageCLEF Wikipedia article on Mauritius.
    The first three predictions are reasonable, the last two are not.
    The third result is
    particularly good given that only two
    sentences mention dodos; for comparison, the object-detection's
    choice began ``(Mauritian Creole people usually known as `Creoles')''.
}
  \label{fig:wiki_predictions}
\end{figure*}

\mparagraph{\diy (new)} We downloaded a sample of 9K Reddit posts made
to the community DIY (``do it yourself''). These posts
\footnote{We
  required at least 25 upvotes per Reddit post to filter out spam and
  low-quality submissions.}
consist of multiple images that users have
taken of the progression of their construction projects, e.g.,
building a rock climbing wall (see Figure~\ref{fig:datasets}). Users
are encouraged to explicitly annotate individual images with
captions,\footnote{
As with \rqa, DIY captions are not
always
grammatical. %
} and, for evaluation, we treat a
caption written alongside a given image as corresponding to a true link.

We adopt the same experimental protocols as in \S \ref{sec:dataone},
but increase the maximum sentence token-length from 20 to 50;
Table~\ref{tab:webdata_results} shows the test-set results.
In general, the algorithms we
introduce again outperform the NoStruct baseline.
In contrast to the \dataOne experiments, AP (slightly) outperformed
the other algorithms.\footnote{This
 holds even when
  varying the number of negatively sampled documents; see the
  \suppmat.
} \diy is the most difficult
among the datasets we consider.%

To see if the algorithms
err
on the same instances,
we again compute the Spearman correlation $\rho$ between test-instance \auc
scores for DC/TK/AP, for $\nnegdoc=10$.
We find greater variation in performance
on \dataTwo
compared to \dataOne data. For example, on \rqa, DC and AP have a
$\rho$ of only $.64$. We also repeat the regression on test-instance
\auc scores introduced in \S\ref{sec:content_spread} with different
results; content generally explains more variance than
spread, e.g., for AP, for \rqa/\diy respectively, only 2/1\% is explained by spread alone, but
18/13\% is explained by spread+content.

\section{Qualitative Exploration}
\label{sec:sec-with-wiki}

To visualize the within-document prediction for document $i$, we
compute $\widehat \simmat_i$ and solve the linear assignment problem
described in \S\ref{sec:section_with_assignment_problem}, taking the
edges with highest selected weights to be the most
confident. Figure~\ref{fig:example_predictions_all} contains example
test predictions (along with $\widehat \simmat_i$)
from the datasets with ground-truth annotation.
 In an effort to
provide representative cases, the selected examples have \auc scores
close to average performance for their corresponding
datasets.

The model mostly succeeds at associating literal objects and their
descriptions: tennis players in \mscoco, castles in \dii, a stapler in \diy, and bacon in a blender
in \rqa.
Errors are often justifiable.
For example, for the \mscoco document,
the chosen caption for a picture of
two people playing baseball accurately
describes the image, despite it having been written for a different image
and thus counting as an error in our quantitative evaluation.
Similarly, for \rqa, a
container of maple syrup is associated with a caption mentioning
``syrup'', which seems reasonable even though the recipe's author did not link that
image/sentence.

In other cases, the algorithm struggles with what part of the image to
``pay attention'' to. In the \dii case
(Figure~\ref{fig:example_prediction_dii}), the algorithm erroneously
(but arguably justifiably) decides to assign a caption about a bride, groom,
and a car to a picture of the couple, instead of to a picture of a
vehicle.

For more difficult datasets like \sis
(Figure~\ref{fig:example_prediction_sis}), the algorithm struggles
with ambiguity. For 2/5 sentences that refer to literal
objects/actions (soup cans/laughter), the algorithm works
well. The remaining 3 captions are general musings about
working at a grocery store that could be matched to any of
the three remaining images depicting grocery store aisles.
\diy is similarly difficult, as
many images/sentences could reasonably be
assigned to each other.

\mparagraph{\wiki} We also constructed a dataset from English
sentence-tokenized Wikipedia articles (not including captions) and their associated images from ImageCLEF2010 \cite{popescu2010overview}. In contrast to \rqa and \diy, there are no explicit connections between individual images and individual sentences,
so we cannot compute \auc or precision, but this corpus represents
an important \dataTwo setting.
We follow the same experimental settings as in \S \ref{sec:dataone} at
training time, but instead of using pre-extracted
features, we fine-tune the
vision model's parameters.\footnote{In comparable settings, fine-tuning the vision CNN
  yields $\approx 20\%$ better performance in terms of the loss in
  Equation~\ref{eq:loss_function} computed over the validation/test
  sets. For memory reasons, we switched from DenseNet169 to
  NASNetSmall \cite{zoph2018learning}; additional details
  are
  in the \suppmat.}
Examining the predictions of the AP+fine-tuned CNN
model trained on \wiki shows many of the model's predictions to be
reasonable.  Figure~\ref{fig:wiki_predictions} shows the model's 5 most
confident predictions on the 100-sentence Wikipedia article about
Mauritius,
chosen
for its high image/text
spread.

\section{Additional related work}

Our similarity functions
are inspired by work in aligning
image fragments, such as object bounding boxes, with portions of
sentences without explicit labels
\cite{karpathy2014deep,karpathy2015deep,jiang2015deep,rohrbach2016grounding,datta2019align2ground};
similar tasks have been addressed in supervised
\cite{plummer2015flickr30k} and semi-supervised \cite{rohrbach2016grounding} settings.
Our models
operate at the larger
granularity
of entire images/sentences.
Integer programs like AP have been used to align visual
and textual content in videos, e.g., \newcite{bojanowski2015weakly}

Prior work has addressed
the task of identifying objects in single images that are referred to
by natural language descriptions
\cite{mitchell2010natural,mitchell2013typicality,kazemzadeh2014referitgame,karpathy2014deep,plummer2015flickr30k,hu2016natural,rohrbach2016grounding,nagaraja2016modeling,hu2016segmentation,yu2016modeling,peyre2017weakly,margffoy2018dynamic}. In
general,
a supervised approach is taken
\citep{mao2016generation,krishna2017visual,johnson2017clevr}.

Related tasks involving multi-image/multi-sentence data include:
generating captions/stories for image streams or videos
\cite{park2015expressing,huang2016visual,shin2016beyond,liu2017let},
sorting aligned (image, caption) pairs into stories
\cite{agrawal2016sort}, image/textual cloze tasks
\cite{iyyer2017amazing,yagcioglu2018recipeqa}, augmentation of
Wikipedia articles with 3D models \cite{russell20133d},
question-answering \cite{kembhavi2017you}, and aligning books
with their film adaptations \cite{zhu2015aligning}; these tasks are
usually supervised, or rely on a search engine.

\section{Conclusion and Future Directions}

We have demonstrated that a family of models for learning fine-grained
image-sentence links \emph{within documents} can produce good
test-time results even if only given access to document-level
co-occurrence at training time.
Future work could incorporate better models of sequence within
document context \citep{kim2015ranking,alikhani2018exploring}. While
using structured loss functions improved performance,
image and sentence \emph{representations} themselves have no awareness
of neighboring images/sentences; this information should
prove useful if modeled appropriately.\footnote{Attempts to
  incorporate document context information
  by passing the word-level RNN's
  output through a
  sentence-level RNN \cite{li2015hierarchical,yang2016hierarchical}
  did not improve performance.}

\newcommand{\firstname}[2]{#1} %

\mparagraph{Acknowledgments}
We thank
\firstname{Yoav}{Y.} Artzi,
\firstname{Cristian}{C.} Danescu-Niculescu-Mizil,
\firstname{Jon}{J.} Kleinberg,
\firstname{Vlad}{V.} Niculae,
\firstname{Justine}{J.} Zhang,
the Cornell NLP seminar,
the reviewers,
and
\firstname{Ondrej}{O.} Linda,
\firstname{Randy}{R.} Puttick,
\firstname{Ramin}{R.} Mehran,
and
\firstname{Grant}{G.} Long
of Zillow
for helpful comments.
We additionally thank
the NVidia Corporation for the GPUs used in this study.
This material is
supported by the U.S. National Science
Foundation under grants BIGDATA SES-1741441,
1526155, 1652536, and the Alfred P. Sloan Foundation.
Any opinions, findings, and conclusions or recommendations expressed in this
material are those of the authors and do not necessarily
reflect the views of the sponsors. \par

\bibliography{refs}
\bibliographystyle{acl_natbib}

\end{document}


\maketitle

\section{Data preprocessing details}

\mparagraph{\mscoco} We downloaded the train/val 2017 images, and the train/val annotations
from 2014 and 2017 from the MSCOCO website (but create our own
training and validation splits). Then, we randomly designate half
of the images as ``true'' images (which will eventually be paired with
their true captions in documents) and half of the images as ``fake''
images, which will not be paired with their true captions in
documents. Then, we randomly group all true images into groups of five,
and all fake images into groups of five. Then, we pair each real-image
set with a fake image set, and divide the resulting groups of 10
images into train/validation/test splits. Then, for each of the
training/validation/testing document sets independently, for each
document, we create (usually) 5 true versions of each document (for
testing and validation, we only sample a single version of each
document, and do not consider the alternate true captions provided by
MSCOCO) because (in general) each MSCOCO image comes with 5 caption
annotations. For each of these true versions, we randomly sample
captions from a pool of all captions written on all images not in that
document (but from the train/validation/test pools independently, so that
there is no overlap between these sets, except in cases where captions
happen to be identical). Then, we shuffle the sampled captions for
each version. The result is 4968/1655/1655 train/validation/test documents,
but each training ``document'' generally consists of 5 versions
because MSCOCO images generally come with 5 captions each.

\mparagraph{\dii/\sis} We downloaded the \dii/\sis train/validation/test
splits along with all images from the
Visual Storytelling Dataset website;\footnote{
\url{http://visionandlanguage.net/VIST/}} we
preserve these splits for our train/validation/test sets. DII stories have
multiple annotations per fixed image set, whereas SIS stories
have multiple annotations per Flickr album, as human annotators were
allowed to select images for their story from all the images within an
album. We discard any story with any invalid or missing image (the FAQ
page on the data download website mentions that images may be missing
because users deleted them).

\mparagraph{\diihard} We augmented the documents from \dii with 45
distractor captions (i.e., captions that were not written about any of
the images in the document) selected uniformly at random. To preserve
train/validation/test splits, we limit these uniform selections to
within-split samples, i.e., training document distractor captions are
sampled only from training documents.

\mparagraph{\rqa} We download the train and validation questions
(29.6K/3.5K) and extract the ``context'' of each question, which
consists of a list of recipe steps and their associated images;
without filtering, there are 8.1K unique recipes in the training set,
and 983 unique recipes in the validation data. We also download the
training/validation images provided. We treat the provided validation
split as the test data.

We concatenate the title and the body of the step
(separating them with a space).
We discard recipe steps that do not contain any
tokens, and discard recipes for which there are no images that
correspond to steps (e.g., if the only steps for which there were
images contained empty text). Then, we reserve training recipes to act
as our validation split. Then, we discard all recipes with fewer than 2
images/recipe steps. The result is 6502/946/878
training/validation/test recipes, with 69K total images. The sizes of
the documents are: mean/median/max number of images: 11/8/93; and
mean/median/max number of sentences: 7/6/20.

\mparagraph{\diy} We downloaded all the submissions on pushshift.io's
files page from Jan.~2013-Oct.~2018. We looped over all of them and
found the ones available made to the subreddit ``DIY,'' for 241K
posts. Then, we discard posts with score less than 25. While the
semantics of the Reddit ``score'' field have changed over time,\footnote{
Other confounding factors: Reddit has become more popular over time,
DIY has likely changed in popularity, etc.} we intend for this
filtration step to act as a basic spam filter. We only consider link
submissions to imgur urls with ``/a/'' in the url, indicating that the
imgur link is an album, rather than a single image. We then scrape the
associated imgur album page and search for all ``div'' html fields
that are ``post-image-container,'' and extract both the image
associated with that field and its associated caption, if it's not
empty; users may leave image captions empty, but may not upload a
caption without an associated image. We ignore imgur albums with no
``post-image-container'' fields. There are 13K documents after this
step. We attempt to scrape all images for these documents, discarding
gifs and invalid images for simplicity, resuling in
295K images.

Next, we search for any image duplicates using \texttt{findimagedupes}
(\url{https://gitlab.com/opennota/findimagedupes}) with a neighbor threshold
of 3. We discard any documents with any duplicate images. Then, we
discard all documents without at least 2 image captions with at least
5 tokens, and discard documents without at least 2 valid
images. Because a small number of documents are quite long, we discard
documents with more than 40 images or more than 40
captions.\footnote{At this step, its possible for there to be more
  captions than images in a document, e.g., because we discard
  animated gifs that may have been associated with captions.} We split
the remaining documents into 6.8K/1K/1K train/validation/test
documents. Between these documents, there are 154K unique images. The
sizes of the documents are: mean/median/max number of images:
17.4/16.0/40; mean/median/max number of sentences: 16.4/15.0/40.

\mparagraph{\wiki} We downloaded the English-language subset of the ImageClef 2011 Wikipedia retrieval
data as a starting point (\url{https://www.imageclef.org/wikidata}).
This dataset contains the full text of Wikipedia articles,
alongside a list of images in each
article. We then stripped out wiki formatting, and used Spacy's
(\url{https://spacy.io/}) English-sentence tokenizer to split
documents into sentences (the resulting sentence tokenization is
imperfect, but sufficient). We keep only the first 100 identified
sentences in a document. We discarded documents with fewer than 10
sentences, and documents with fewer than 3 images. The result is 16K
articles, for which we used a 14K/1K/1K train/validation/test split. For the results
discussed in the paper, we explore same-document predictions on
training documents using a model checkpoint with low validation
error. The sizes of the documents are: mean/median/max number of images:
6/5/108, mean/median/max number of sentences: 72/86/100.

\mparagraph{Download}
All datasets are available for download: \url{www.cs.cornell.edu/\~jhessel/multiretrieval/multiretrieval.html}

\begin{figure*}[t!]
  \centering
  \begin{subfigure}[t]{0.5\linewidth}
    \centering
    \includegraphics[width=.75\linewidth]{figures/dynamics/labels_version_4/mscoco-crop.pdf}
    \caption{\mscoco}
    \label{fig:sup_mscoco_dynamics}
  \end{subfigure}~
  \begin{subfigure}[t]{0.5\linewidth}
    \centering
    \includegraphics[width=.75\linewidth]{figures/dynamics/labels_version_4/dii-crop.pdf}
    \caption{\dii}
    \label{fig:sup_dii_dynamics}
  \end{subfigure}
  \begin{subfigure}[t]{0.5\linewidth}
    \centering
    \includegraphics[width=.75\linewidth]{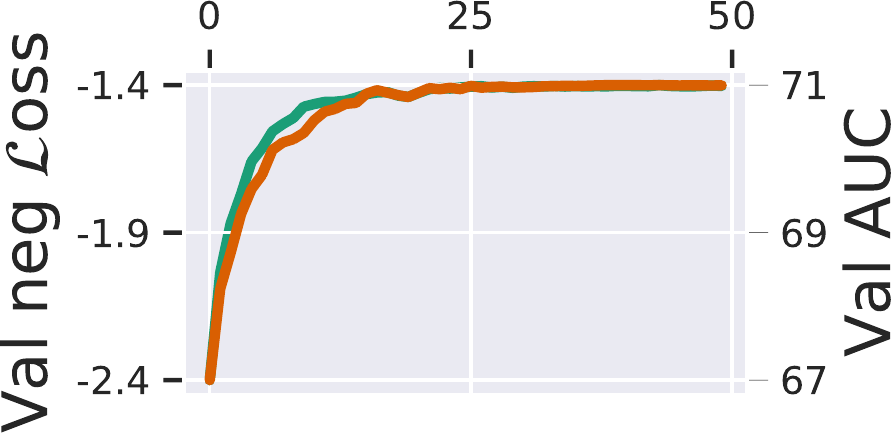}
    \caption{\sis}
    \label{fig:sup_sis_dynamics}
  \end{subfigure}~
  \begin{subfigure}[t]{0.5\linewidth}
    \centering
    \includegraphics[width=.75\linewidth]{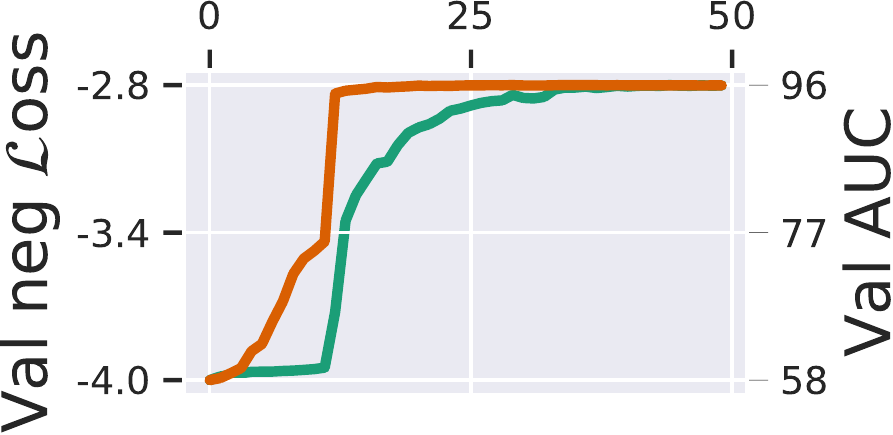}
    \caption{\diihard}
    \label{fig:supp_diir_dynamics}
  \end{subfigure}
  \begin{subfigure}[t]{0.5\linewidth}
    \centering
    \includegraphics[width=.75\linewidth]{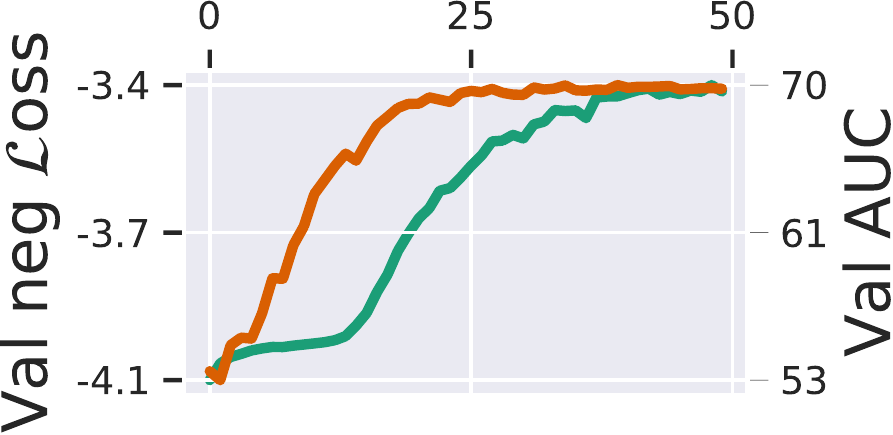}
    \caption{\rqa}
    \label{fig:supp_rqa_dynamics}
  \end{subfigure}~
  \begin{subfigure}[t]{0.5\linewidth}
    \centering
    \includegraphics[width=.75\linewidth]{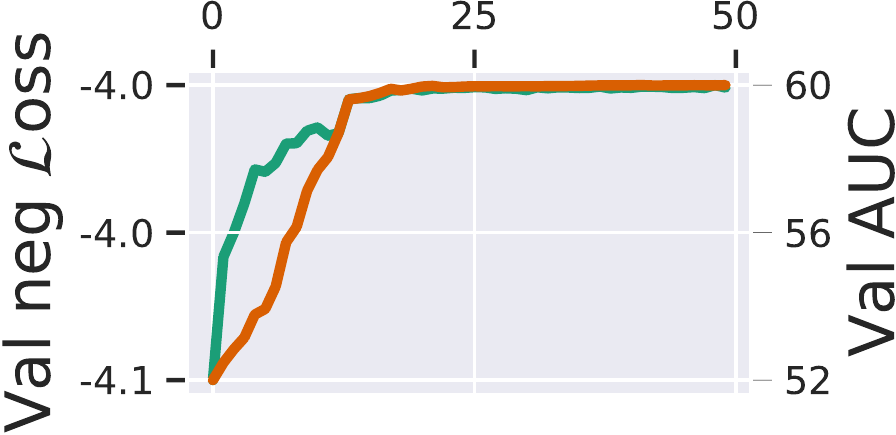}
    \caption{\diy}
    \label{fig:supp_diy_dynamics}
  \end{subfigure}
  \caption{Inter-document objective (AP, $\nnegdoc=10$, hard negative
    mining) and intra-document \auc during 50 epochs of training for
    all datasets we consider with ground-truth, intra-document
    annotations. While there are some interesting discontinuities,
    e.g., in \diihard's training curves, in general, for a fixed
    neural architecture/similarity function, better retrieval
    performance, as measured by the negative-loss computed over the
    validation set, equates to better intra-document performance, as
    measured by \auc.}
  \label{fig:supp_outcomes}
\end{figure*}

\section{\wiki Fine-tuning Details}

We experiment with fine-tuning the parameters of our image model for
the \dataTwo data, as an alternative to extracting features from a pretrained
network. However, given that hundreds of images and sentences need
to fit in GPU memory for each batch (we worked with a single
GPU with 12GB of RAM), we needed to switch our CNN from
DenseNet169 to one with a smaller memory footprint; we chose
NASNetSmall.  But even so, we still require a word-embedding
matrix and a 1024-dimensional GRU in memory. Hence,
additionally, at training time, for documents with more
than 10 images/sentences, we randomly downsample images/sentences to a
set of 10 (though at validation and test time, longer documents are
kept intact).
This
subsampling process
ensures that at most 110 images are in GPU memory at a time (for 10
negative samples per positive sample). When training the CNN, we also
perform random data augmentation to help regularize. We first resize
images to 256 by 256, and, at training time, perform the following
data augmentation: random horizontal flipping, up to 20 degree random
image rotation, and a random crop to 224 by 224. At validation/test
time, we use a center crop (with no rotations or flips).

We trained models with AP using fixed, NASNetSmall pre-extracted
features, and compared those models to ones where we fine-tuned the
additional 5M CNN parameters. The resulting \emph{test}
\newcommand{\negloss}{-{\cal L}}
\auc/negative-loss ($\negloss$) values are:

\begin{center}
  {\footnotesize
    \begin{tabular}{lc@{\hspace{.1cm}}cc@{\hspace{.1cm}}cc@{\hspace{.1cm}}c}
      & \multicolumn{2}{c}{\rqa} & \multicolumn{2}{c}{\diy} & \multicolumn{2}{c}{\wiki} \\
      & \auc & $\negloss$ & \auc & $\negloss$ & \auc & $\negloss$ \\
      \midrule
      Fixed CNN & \textbf{67.6} & \textbf{-.37} & \textbf{60.9} & \textbf{-.37} & N/A & -.26 \\
      Finetuned CNN & 65.7 & -.40 & 57.9 & -.39 & N/A & \textbf{-.21} \\
    \end{tabular}
  }
\end{center}
Thus, we did not observe intra-document performance increases with
fine-tuning for \diy and \rqa for the experiment settings we
consider. However,
on \wiki, for negative-training-loss (the only metric we can compute on this no-ground-truth
dataset),  fine-tuning performed better. \footnote{Fine-tuning NASNetSmall also
beat using DenseNet169 extracted features.
} %
Since Figure~\ref{fig:supp_outcomes} demonstrates that, for a fixed architecture
and for datasets where \auc can be computed, \auc and (the negative of) training
loss rise together, we expect that fine-tuning is beneficial for \wiki.

\section{Additional Results}

Tables containing our full results are given in Tables~\ref{tab:ground_truth_dataone_res_20},~\ref{tab:ground_truth_dataone_res_30},~\ref{tab:ground_truth_res_20}, and~\ref{tab:ground_truth_res_30}.
Compared to
the results presented in the paper, here we explicitly compare
additional hyperparameter configurations. Specifically: we show
results for $b=10,20,30$ negative samples (the main paper just shows
$b=10$) and compare using hard negative mining vs.~not using hard
negatives (the main paper just shows hard negative mining results,
e.g., ``AP+hard neg'' in these tables is the same as the ``AP''
described in the main paper). In general, hard negative mining
improves performance, and the number of negative samples doesn't
greatly affect performance in the range we examined.

\begin{table*}
  \small
  \centering
  
\begin{tabular}{l|cc|cc|cc|cc}
\toprule
&\multicolumn{2}{|c|}{\mscoco}&\multicolumn{2}{|c|}{\dii}&\multicolumn{2}{|c|}{\sis}&\multicolumn{2}{|c}{\diihard}\\
& \auc & \patone/\patfive & \auc & \patone/\patfive & \auc & \patone/\patfive & \auc & \patone/\patfive \\
 \midrule
Random& 49.7 & 5.0/4.6 & 49.4 & 19.5/19.2 & 50.0 & 19.4/19.7 & 50.0 & 2.0/2.0  \\
Obj Detect& 89.5 & 67.7/45.9 & 65.3 & 50.2/35.2 & 58.4 & 40.8/28.6 & 76.9 & 25.7/17.5  \\
NoStruct& 88.3 & 53.4/35.8 & 76.6 & 60.4/46.2 & 64.9 & 43.3/33.8 & 84.2 & 21.4/15.6  \\
NoStruct+ hard neg& 51.8 & 8.3/5.9 & 75.9 & 63.0/45.0 & 63.3 & 45.1/31.9 & 51.9 & 4.3/3.1  \\
\midrule
DC& 98.8 & 92.0/78.6 & 81.8 & 69.1/53.7 & 68.0 & 49.7/37.6 & 93.8 & 58.3/40.1  \\
DC+ hard neg& 98.9 & 93.1/79.9 & 82.9 & 71.9/55.7 & 68.8 & 52.2/38.7 & 95.0 & 65.2/44.9  \\
\midrule
TK& 98.8 & 92.1/78.6 & 81.8 & 69.6/53.8 & 68.0 & 49.7/37.6 & 94.4 & 60.2/42.2  \\
TK+ hard neg& 98.9 & 93.9/80.0 & 82.8 & 71.5/55.7 & 68.8 & 51.8/38.5 & 95.2 & 65.2/45.3  \\
TK+ hard neg+ $\frac{1}{2}k$& 99.0 & 95.0/81.4 & 81.9 & 71.4/54.5 & 67.6 & 51.5/37.8 & 94.7 & 64.5/43.4  \\
\midrule
AP& 98.5 & 87.6/75.3 & 81.7 & 68.3/53.5 & 67.3 & 47.1/36.6 & 93.5 & 58.3/39.7  \\
AP+ hard neg& 98.7 & 91.1/77.9 & 82.6 & 70.7/55.0 & 68.6 & 50.6/38.3 & 95.4 & 65.4/45.5  \\
AP+ hard neg+ $\frac{1}{2}k$& 98.9 & 94.1/80.7 & 81.5 & 72.2/54.2 & 67.4 & 51.9/37.7 & 94.6 & 64.7/43.7  \\
\bottomrule
\end{tabular}

  \caption{Results for \dataOne data with ground-truth annotation with
    $\nnegdoc=20$ negative samples.}
  \label{tab:ground_truth_dataone_res_20}
\end{table*}

\begin{table*}
  \small
  \centering
  
\begin{tabular}{l|cc|cc|cc|cc}
\toprule
&\multicolumn{2}{|c|}{\mscoco}&\multicolumn{2}{|c|}{\dii}&\multicolumn{2}{|c|}{\sis}&\multicolumn{2}{|c}{\diihard}\\
& \auc & \patone/\patfive & \auc & \patone/\patfive & \auc & \patone/\patfive & \auc & \patone/\patfive \\
 \midrule
Random& 49.7 & 5.0/4.6 & 49.4 & 19.5/19.2 & 50.0 & 19.4/19.7 & 50.0 & 2.0/2.0  \\
Obj Detect& 89.5 & 67.7/45.9 & 65.3 & 50.2/35.2 & 58.4 & 40.8/28.6 & 76.9 & 25.7/17.5  \\
NoStruct& 87.5 & 50.8/34.7 & 76.6 & 59.9/46.2 & 64.9 & 43.4/33.7 & 84.1 & 21.3/15.6  \\
NoStruct+ hard neg& 52.0 & 10.3/6.0 & 75.9 & 63.0/45.0 & 63.0 & 44.5/31.5 & 51.8 & 4.0/2.9  \\
\midrule
DC& 98.8 & 92.0/78.7 & 82.2 & 70.5/54.6 & 68.0 & 49.7/37.7 & 93.9 & 58.6/40.3  \\
DC+ hard neg& 98.9 & 93.4/79.9 & 82.8 & 71.3/55.5 & 68.8 & 52.1/38.6 & 95.0 & 63.8/44.5  \\
\midrule
TK& 98.8 & 91.6/78.7 & 81.8 & 69.5/53.9 & 68.0 & 49.9/37.7 & 94.4 & 60.5/42.4  \\
TK+ hard neg& 98.9 & 93.3/80.0 & 82.8 & 71.4/55.7 & 68.8 & 51.0/38.6 & 95.2 & 65.3/45.7  \\
TK+ hard neg+ $\frac{1}{2}k$& 99.0 & 95.2/81.5 & 82.1 & 73.1/55.1 & 67.7 & 51.9/37.8 & 94.7 & 64.2/43.6  \\
\midrule
AP& 98.5 & 87.3/75.4 & 81.7 & 67.7/53.4 & 67.3 & 47.1/36.6 & 93.4 & 57.2/39.8  \\
AP+ hard neg& 98.7 & 91.2/78.0 & 82.6 & 71.1/55.0 & 68.5 & 50.3/38.2 & 95.3 & 65.3/45.6  \\
AP+ hard neg+ $\frac{1}{2}k$& 98.9 & 94.1/80.5 & 81.6 & 72.8/54.4 & 67.4 & 51.8/37.8 & 94.4 & 64.3/43.2  \\
\bottomrule
\end{tabular}

  \caption{Results for \dataOne data with
    $\nnegdoc=30$ negative samples.}
  \label{tab:ground_truth_dataone_res_30}
\end{table*}

\begin{table*}
  \small
  \centering
  
\begin{tabular}{l|cc|cc|}
\toprule
&\multicolumn{2}{|c|}{\rqa}&\multicolumn{2}{|c}{\diy}\\
& \auc & \patone/\patfive & \auc & \patone/\patfive \\
 \midrule
Random& 49.4 & 17.8/16.7 & 49.8 & 6.3/6.8  \\
Obj Detect& 58.7 & 25.1/21.5 & 53.4 & 17.9/11.8  \\
NoStruct& 60.5 & 34.3/26.8 & 56.9 & 13.8/12.2  \\
NoStruct+ hard neg& 60.1 & 35.0/26.7 & 56.3 & 15.0/12.5  \\
\midrule
DC& 67.1 & 43.8/34.9 & 59.5 & 19.3/15.2  \\
DC+ hard neg& 63.4 & 36.6/31.0 & 59.3 & 21.0/16.0  \\
\midrule
TK& 65.2 & 41.6/33.1 & 60.0 & 20.4/15.5  \\
TK+ hard neg& 67.9 & 45.2/36.0 & 60.5 & 20.3/16.2  \\
TK+ hard neg+ $\frac{1}{2}k$& 67.7 & 44.4/35.0 & 56.1 & 14.8/12.0  \\
\midrule
AP& 66.9 & 37.8/34.2 & 59.1 & 16.9/13.9  \\
AP+ hard neg& 69.4 & 45.9/37.8 & 61.9 & 23.3/17.9  \\
AP+ hard neg+ $\frac{1}{2}k$& 68.5 & 44.9/36.4 & 59.6 & 21.7/15.7  \\
\bottomrule
\end{tabular}

  \caption{Results for \dataTwo data with ground-truth annotation with
    $\nnegdoc=20$ negative samples.}
  \label{tab:ground_truth_res_20}
\end{table*}

\begin{table*}
  \small
  \centering
  
\begin{tabular}{l|cc|cc|}
\toprule
&\multicolumn{2}{|c|}{\rqa}&\multicolumn{2}{|c}{\diy}\\
& \auc & \patone/\patfive & \auc & \patone/\patfive \\
 \midrule
Random& 49.4 & 17.8/16.7 & 49.8 & 6.3/6.8  \\
Obj Detect& 58.7 & 25.1/21.5 & 53.4 & 17.9/11.8  \\
NoStruct& 60.4 & 34.5/26.7 & 56.9 & 13.3/11.9  \\
NoStruct+ hard neg& 59.7 & 31.8/27.0 & 55.9 & 14.7/12.4  \\
\midrule
DC& 66.7 & 42.7/34.1 & 59.5 & 18.9/14.7  \\
DC+ hard neg& 63.5 & 37.6/30.6 & 59.4 & 20.8/16.4  \\
\midrule
TK& 65.3 & 41.2/32.8 & 60.1 & 20.0/15.9  \\
TK+ hard neg& 68.0 & 44.0/36.2 & 60.5 & 21.4/16.1  \\
TK+ hard neg+ $\frac{1}{2}k$& 67.8 & 43.2/35.1 & 57.3 & 19.1/13.5  \\
\midrule
AP& 66.5 & 41.0/33.8 & 59.2 & 15.7/14.0  \\
AP+ hard neg& 69.3 & 47.5/37.4 & 61.9 & 24.4/17.8  \\
AP+ hard neg+ $\frac{1}{2}k$& 68.7 & 45.2/36.2 & 59.4 & 22.0/15.7  \\
\bottomrule
\end{tabular}

  \caption{Results for \dataTwo data with
    $\nnegdoc=30$ negative samples.}
  \label{tab:ground_truth_res_30}
\end{table*}